\documentclass[10pt,twocolumn,letterpaper]{article}

\usepackage[pagenumbers]{iccv} % To force page numbers, e.g. for an arXiv version

\definecolor{iccvblue}{rgb}{0.21,0.49,0.74}
\usepackage[pagebackref,breaklinks,colorlinks,allcolors=iccvblue]{hyperref}

\usepackage{xspace}
\usepackage{xcolor}
\usepackage{colortbl}
\usepackage{makecell,multirow}
\usepackage{xfrac}
\usepackage{units}
\usepackage{tikz}

\def\paperID{6858} % *** Enter the Paper ID here
\def\confName{ICCV}
\def\confYear{2025}

\title{Mixpert: Mitigating Multimodal Learning Conflicts with Efficient Mixture-of-Vision-Experts}

\author{
\textbf{Xin He}$^{1*}$\quad
\textbf{Xumeng Han}$^{2,1*}$ \quad
\textbf{Longhui Wei}$^{1\dagger}$ \quad 
\textbf{Lingxi Xie}$^{1}$ \quad 
\textbf{Qi Tian}$^{1}$
\vspace{0.4em}\\
$^1$Huawei Inc. \quad
$^2$University of Chinese Academy of Sciences 
\vspace{0.4em}\\
$^*${\normalsize Equal Contribution} \quad
$^{\dagger}${\normalsize Corresponding Author}
}

\begin{document}
\maketitle
\newcommand{\name}{\text{Mixpert}\xspace}
\begin{abstract}
Multimodal large language models (MLLMs) require a nuanced interpretation of complex image information, typically leveraging a vision encoder to perceive various visual scenarios.
However, relying solely on a single vision encoder to handle diverse task domains proves difficult and inevitably leads to conflicts.
Recent work enhances data perception by directly integrating multiple domain-specific vision encoders, yet this structure adds complexity and limits the potential for joint optimization. In this paper, we introduce \name, an efficient mixture-of-vision-experts architecture that inherits the joint learning advantages from a single vision encoder while being restructured into a multi-expert paradigm for task-specific fine-tuning across different visual tasks.
Additionally, we design a dynamic routing mechanism that allocates input images to the most suitable visual expert. \name effectively alleviates domain conflicts encountered by a single vision encoder in multi-task learning with minimal additional computational cost, making it more efficient than multiple encoders. Furthermore, \name integrates seamlessly into any MLLM, with experimental results demonstrating substantial performance gains across various tasks.
\end{abstract}

\section{Introduction}
\label{sec:intro}

The breakthrough advancements in large language models (LLMs)~\cite{radford2018improving,radford2019language,touvron2023llama,touvron2023llama2, dubey2024llama} have sparked widespread interest in their potential for visual understanding and reasoning, driving researchers to develop models capable of \emph{seeing} the real world. This demand has given rise to the emergence of multimodal large language models (MLLMs)~\cite{liu2023visual,instructblip,zhu2023minigpt,lin2023sphinx, li2024mini, lu2024deepseek, yao2024minicpm, li2023blip}, which typically employ an architecture where images are converted by a visual encoder into a series of visual tokens and input into the LLM alongside text embeddings. 
The visual encoder typically employs models pre-trained on extensive image-text paired data, such as CLIP~\cite{radford2021learning} and SigLIP~\cite{zhai2023sigmoid}, which embed images into a language-consistent feature space.

\begin{figure}
  \centering
  \includegraphics[width=1\columnwidth]{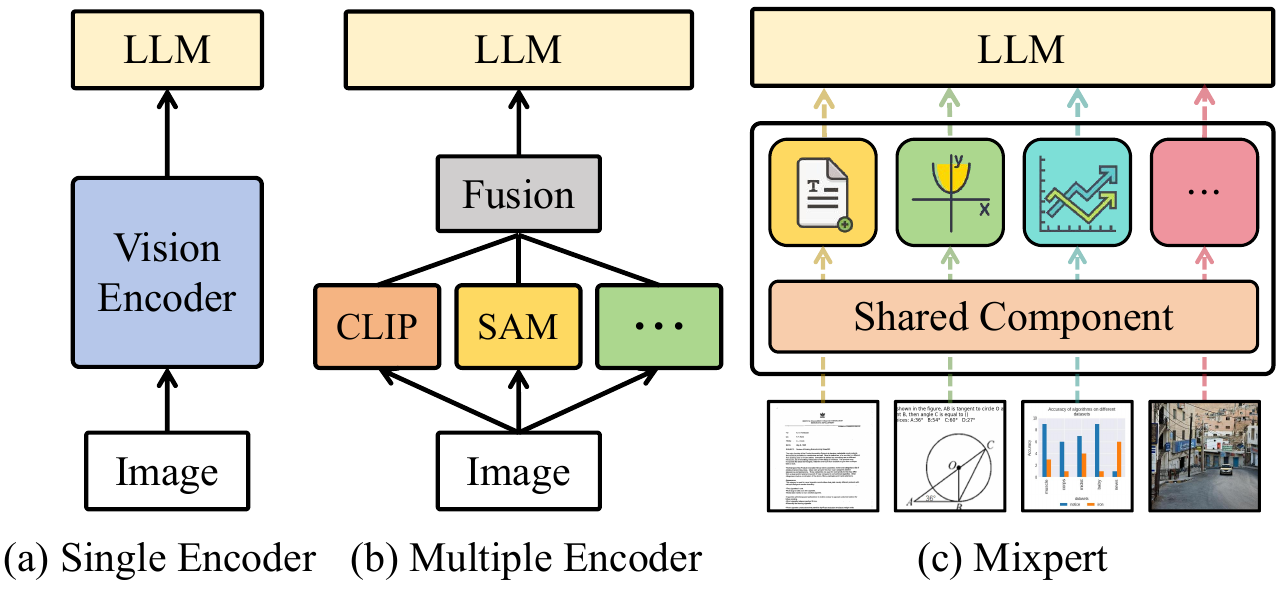}
%   \vspace{-6pt}
  \caption{\textbf{Vision encoder structures.} (a) A single vision encoder handles various task scenarios. (b) Integrating multiple task-specific vision encoders to enhances data perception but introduces additional costs.
  (c)~Our efficient mixture-of-vision-experts framework, \textbf{Mixpert}, assigns each expert to a specific domain. The input is routed to the expert most aligned with its type, incurring minimal computational overhead compared to a single encoder.
}
%   \vspace{-9pt}
  \label{fig:intro}
\end{figure}

Multimodal tasks encompass various visual challenges, each with significantly different image characteristics, making it difficult for a single vision encoder to achieve optimal balance across multiple domains. Recent studies~\cite{lin2023sphinx, he2024incorporating, tong2024cambrian, shi2024eagle} employ multiple vision encoders, illustrated in Fig.~\hyperref[fig:intro]{\ref*{fig:intro}\,(b)}, to extract information from different perspectives, enriching visual content representation and enhancing the perception of various data types. This mode is shown to be effective and helps MLLMs better adapt to diverse visual characteristics and requirements. However, using multiple vision encoders undoubtedly increases computational and deployment burdens. Moreover, the independent nature of multiple visual encoders prevents them from gaining additional benefits through joint optimization with multimodal data, leaving each encoder reliant solely on its pre-trained capabilities. This separated structure limits the ability of vision encoders to fully leverage synergistic advantages in multimodal tasks, thereby reducing the overall optimization potential.

To address the above issues, we draw on the concept of mixture-of-experts (MoE)~\citep{jacobs1991adaptive, jordan1994hierarchical} structure and develop \name to resolve domain conflicts encountered by MLLMs during multi-task joint learning in supervised fine-tuning (SFT).
As show in Fig.~\hyperref[fig:intro]{\ref*{fig:intro}\,(c)}, we partition the vision encoder into a shared component and a multi-expert component.
The shared component corresponds to the shallow layers of the vision encoder and is responsible for extracting features from all types of data, encompassing the foundational information for various visual tasks. The deeper layers and projector are structured into a mixture-of-vision-experts paradigm, where each expert focuses on a specific domain and performs more refined processing tailored to particular tasks.

In contrast to typical MoE models~\cite{gou2023mixture,lin2024moe}, this paper manually defines the domains of experts, with each domain aligning with a specific multimodal data type.
This approach allows each expert to be trained exclusively on its corresponding data type, enabling parameter decoupling for independent learning of different capabilities and mitigating potential conflicts in multitask joint optimization. 
Furthermore, we design a routing network to automatically select the most suitable expert for input images. The router is constructed using a simple two-layer MLP, which classifies the features extracted by the shared component of the vision encoder and assigns the image to the expert with the corresponding confidence.
Since each image is allocated to only one expert, there is no additional computational overhead beyond the router. This design maintains efficient processing while significantly improving performance across tasks from different domains.
The advantages of \name can be summarized as follows:
\begin{itemize}
\item Compared to a single vision encoder, \name effectively alleviates the conflicts arising from multi-task learning in SFT. During the fine-tuning of vision experts, \name enables the flexible injection of task-specific data, obviating concerns around data balance across varied domains.
\item For each image, \name activates only the expert corresponding to the relevant data type. Consequently, the additional activation parameters and computational overhead are minimal, making it more efficient compared to solutions that rely on multiple vision encoders.
\item \name seamlessly integrates into any MLLM. We perform experimental validation on state-of-the-art models such as LLaVA-OV~\cite{li2024llava-ov}, InternVL2~\cite{chen2024far}, and Qwen2-VL~\cite{wang2024qwen2}, with results showing that \name consistently delivers superior performance across all these models.
\end{itemize}
\section{Related Work}
\textbf{Multimodal Large Language Models (MLLMs).} In recent years, MLLMs~\cite{liu2023visual,instructblip,zhu2023minigpt,lin2023sphinx, li2024mini, lu2024deepseek, yao2024minicpm, li2023blip} have garnered sustained attention, primarily attempting to integrate various other modalities into LLMs. Among these, the most widely studied approaches involve encoding visual modalities using visual encoders and then feeding them into LLMs alongside language tokens. For instance, BLIP2~\cite{li2023blip} and InstructBLIP~\cite{instructblip} employ a Q-Former structure to compress visual tokens extracted from CLIP-like encoders, while the LLaVA series~\cite{liu2023visual, liu2024improved, liu2024llava, li2024llava-ov} utilize a simple MLP for projecting visual tokens into LLM embedding space. The Qwen-VL series~\cite{bai2023qwen, wang2024qwen2} and InternVL series~\cite{chen2024internvl, chen2024far} utilize a pixel shuffle operation to reduce the number of visual tokens and then simply utilize MLP layers to project these tokens. 
In addition to methods that utilize a single vision encoder, there are also approaches that attempt to enhance the performance of MLLM by employing multiple visual encoders.
SPHINX~\cite{lin2023sphinx}, IVE~\cite{he2024incorporating}, and Cambrian-1~\cite{tong2024cambrian} have been proposed to incorporate multiple visual encoders to enhance the overall visual perception capability. Despite our approach also integrating multiple visual experts, it differs from the aforementioned methods by proposing to improve the perception capability for different types of visual inputs with minimal additional computation cost.

\noindent
\textbf{Mixture-of-Experts (MoE).} The
MoE~\cite{jacobs1991adaptive, jordan1994hierarchical} models consist of multiple expert networks designed to handle different types of input data, with a router selecting the most suitable experts to process each sample. As an efficient architecture, MoE models~\cite{lepikhin2020gshard, shazeer2017sparsely, zhou2022mixture,han2024vimoe} have garnered widespread attention.
To optimize resource utilization, many models~\citep{switch_transformers,Gshard,Glam} in the field of natural language processing (NLP) adopt sparse MoE architectures to handle large-scale tasks.
In the quest to design more efficient MLLMs, MoCLE~\cite{gou2023mixture} and MoE-LLaVA~\cite{lin2024moe} implement the MoE design in LLM with fewer parameter increases,  but have achieved comparable performance to those with larger LLMs.
 In contrast to previous methods~\cite{gou2023mixture, lin2024moe, liu2024sphinx} which primarily borrowed schemes from NLP and design the MoE architectures in LLM, we have developed a more efficient MoE structure. Our approach involves designing the MoE in the visual encoder and projector layer to resolve domain conflicts of input images, and we achieve a notable reduction in parameters compared to earlier methods.
\section{Methodology}

\begin{figure*}
    \centering
    \includegraphics[width=\linewidth]{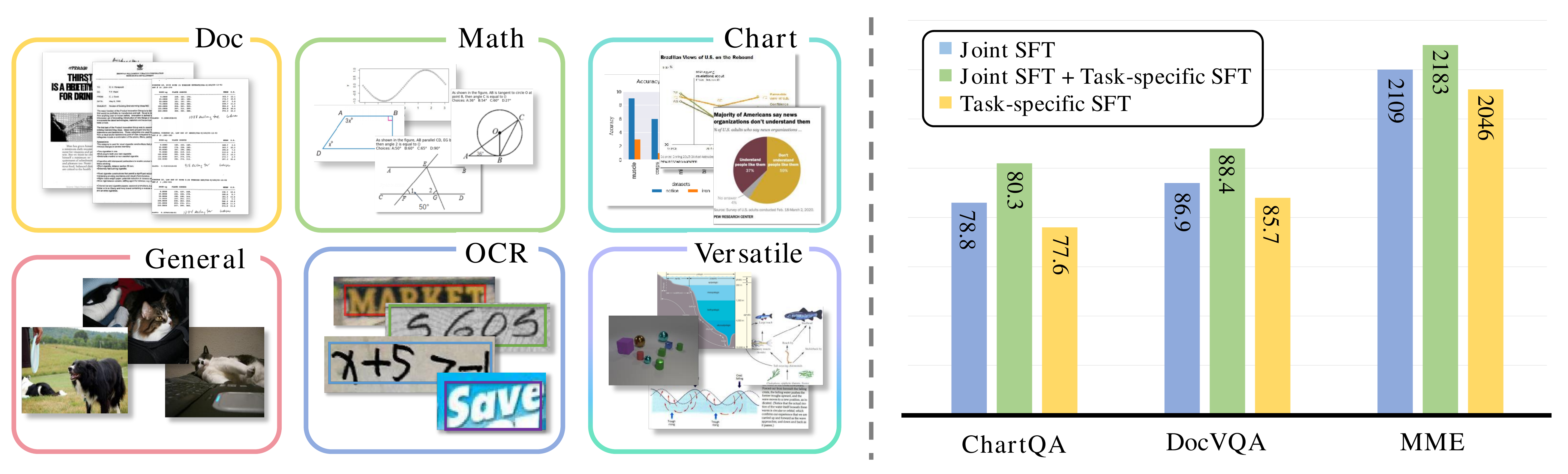}
    \caption{\textbf{(Left)} Multimodal data encompasses a variety of types, with images differing widely in content and structure. \textbf{(Right)}~Building upon the joint SFT model, further domain-specific specialization can enhance performance on the respective task. This also demonstrates that joint optimization, due to balancing multiple objectives, limits the ability to achieve optimal performance in each individual domain. However, SFT on a single task alone does not yield satisfactory results, underscoring the necessity of joint learning for leveraging common knowledge across diverse data.}
    \label{fig:motivation}
\end{figure*}

\subsection{Preliminary}
\label{sec:preliminary}
Multimodal large language models (MLLMs)~\cite{liu2023visual,li2024llava-ov} typically comprise three core components: \textbf{\emph{(\romannumeral 1)}} a vision encoder $g(\cdot)$, which encodes the input image $\mathbf{X_v}$ into a visual embedding $\mathbf{Z_v}=g(\mathbf{X_v})$; \textbf{\emph{(\romannumeral 2)}} a projector $p(\cdot)$, usually a multi-layer perceptron (MLP), which projects the image features to the word embedding space $\mathbf{H_v}=p(\mathbf{Z_v})$; and \textbf{\emph{(\romannumeral 3)}} a large language model (LLM), which generates the response $\mathbf{X_a}$ based on the visual embedding $\mathbf{H_v}$ and the tokenized language instruction $\mathbf{H_q}$.

The training of MLLMs primarily involves pre-training and supervised fine-tuning (SFT). Taking the advanced LLaVA-OV~\cite{li2024llava-ov} as an example, its pre-training consists of two sub-stages: language-image alignment \emph{(Stage-1)}, which is trained on 558K data, and high-quality knowledge learning \emph{(Stage-1.5)}, where 4M data is used to inject additional knowledge. The SFT \emph{(Stage-2)} in LLaVA-OV includes two modes: single-image (SI) training with 3.2M data and OneVision version using a mixture of video, single-image, and multi-image data. This paper only considers the single-image mode, and all subsequent mentions of LLaVA-OV refer to LLaVA-OV-7B (SI).

\subsection{Motivation}
\noindent
\textbf{Domain Conflicts in Multi-task Learning.}
MLLMs encounter diverse visual challenges in real-world scenarios, with each task exhibiting distinct image characteristics. As shown in Fig.~\hyperref[fig:motivation]{\ref*{fig:motivation}\,(Left)}, tasks vary significantly in terms of image content and focal details. This diversity requires the visual encoder to adapt to each type, yet it limits the ability to fully realize its potential across multiple domains. When the model attempts to learning multiple tasks concurrently, inter-task competition emerges, constraining the attainment of potential performance for each individual task. We experimentally verify the existence of domain conflicts on LLaVA-OV~\cite{li2024llava-ov}. Specifically, starting from the joint SFT \emph{(Stage-2)} checkpoint, we further fine-tune the model separately on the \emph{Chart}, \emph{Doc}, and \emph{General} domains using corresponding visual instruction tuning data provided by LLaVA-OV~\cite{li2024llava-ov} (more details are provided in Sec.~\ref{sec:4_1}). Here, we fine-tune all the parameters of the MLLM. As shown in Fig.~\hyperref[fig:motivation]{\ref*{fig:motivation}\,(Right)}, the task-specific fine-tuned model outperforms the original across all three domains, indicating that multi-task joint learning somewhat restricts optimal performance in each domain.

\noindent
\textbf{Advantages of Joint Optimization.}
While we highlight the domain conflicts in multi-task learning, it does not suggest that joint optimization is inherently ineffective. On the contrary, joint SFT leverages a broader range of data, thereby fostering the acquisition of foundational capabilities and common knowledge.
In Fig.~\hyperref[fig:motivation]{\ref*{fig:motivation}\,(Right)}, we report the results of non-joint training, \ie, using the pre-trained LLaVA-OV \emph{(Stage-1.5)} weights and fine-tuning on domain-specific data only.
The results show that joint optimization significantly outperforms task-specific SFT, indicating its advantages in integrating domain knowledge, enhancing single-task performance, and fostering inter-task synergy.

\noindent
\textbf{Discussion.}
From the above experiments, we conclude that joint optimization enables the model to acquire robust and comprehensive foundational capabilities by exposing it to a broader spectrum of data across various scenarios. On the other hand, task-specific SFT allows the model to specialize and overcome the limitations imposed by domain conflicts. We attempt to leverage the strengths of both joint optimization and task-specific SFT, enabling the vision encoder to assimilate shared knowledge while simultaneously responding to the distinct demands of each domain.

\begin{figure*}[t]
  \centering
  \includegraphics[width=0.97\textwidth]{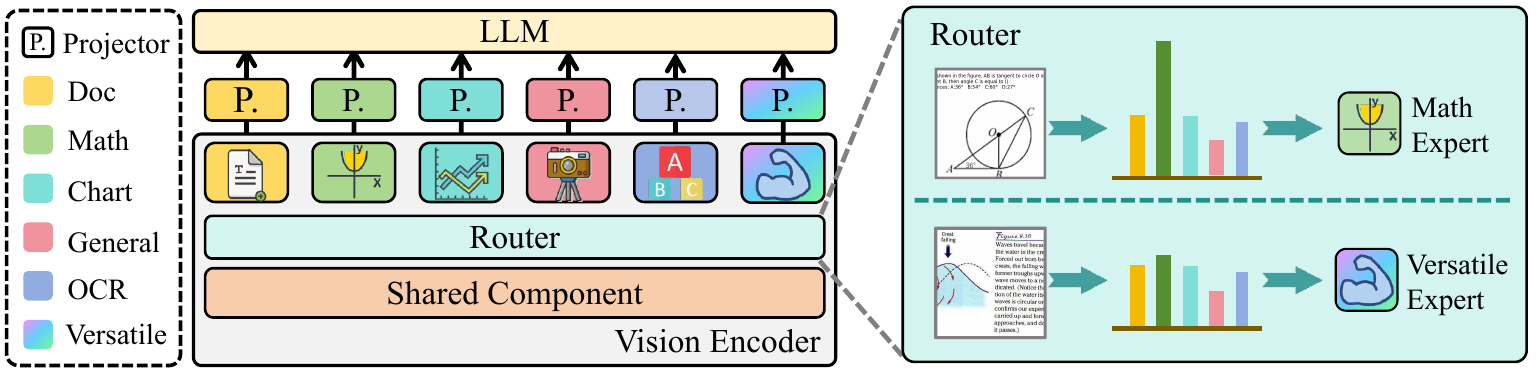}
%   \vspace{-7pt}
\caption{\textbf{Architecture of \name.} We build a multi-expert component within the vision encoder, with each expert responsible for a specific task. This allows for specialization even when there are significant differences between domains, effectively mitigating potential conflicts. For images with ambiguous types or multiple characteristics, the router struggles to make a decisive prediction, thus directing them to the versatile expert for handling.
 }
%   \vspace{-11pt}
  \label{fig:framework}
\end{figure*}

\subsection{\name: Efficient Mixture-of-Vision-Experts}
\label{sec:3_3}
\noindent
\textbf{Vision Expert Task Allocation.}
We design \name, as illustrated in Fig.~\ref{fig:framework}, to leverage the benefits of multi-task joint learning while alleviating domain conflicts. Based on the MLLM that has undergone joint SFT, \name incorporates an MoE~\cite{jacobs1991adaptive, jordan1994hierarchical} structure within the vision encoder, employing multiple experts to address tasks across various domains and enabling flexible adaptation to the demands of diverse scenarios. Specifically, we partition the vision encoder (including the projector) into two parts: the shallow layers serve as a shared component $g_s(\cdot)$ to extract common information, while the deeper layers and projector are structured as an MoE, enabling specialized processing for different data through a divide-and-conquer manner.

In contrast to typical MoE models~\cite{gou2023mixture,lin2024moe}, we do not rely on the model to learn how to partition the data and allocate inputs to the experts (the core challenge while optimizing MoE). Instead, we manually predefine the data types and assign each type to a dedicated expert.
This design delineates the task allocation and specialization for each expert, thereby enhancing the interpretability of their respective roles. We categorize multimodal data into five primary types: chart analysis~\cite{masry2022chartqa, kantharaj2022chart, kafle2018dvqa}, document recognition~\cite{mathew2021docvqa, tanaka2021visualmrc}, mathematical reasoning~\cite{gao2023g, kazemi2023geomverse}, OCR~\cite{yuan2022syntax, mishraICDAR19}, and general understanding~\cite{schwenk2022okvqa, chen2023sharegpt4v}, assigning a corresponding expert $e_i(\cdot)$ $(i=1,2,3,4,5)$ to each data type.

\noindent
\textbf{Disentangled Vision Expert Tuning.}
We propose a simple yet effective disentangled vision expert tuning strategy to further enhance the specialization of each expert within its domain. We freeze all common modules, including the LLM and the shared component of the vision encoder, initialize the experts with the weights from joint SFT, and fine-tune them using domain-specific data.
This approach maintains the robustness of the common modules while empowering each expert to achieve in-depth optimization within its specialized domain, thereby elevating the performance across various tasks.
Another advantage of disentangled tuning is that it eliminates the need to consider the data distribution across domains, allowing each expert to learn directly from the data corresponding to its specific task.

\noindent
\textbf{Versatile Expert.}
In practical applications, MLLMs are tasked with addressing an infinite variety of scenarios, while the predefined data types are limited, preventing a precise and exhaustive categorization of all images. Furthermore, as illustrated in Fig.~\hyperref[fig:motivation]{\ref*{fig:motivation}\,(Left)}, some images encompass multiple domains, complicating a singular categorization based solely on visual content.
To mitigate these issues, we introduce a versatile expert $e_v(\cdot)$, which reuses the weights from the original joint SFT, enabling flexible adaptation to various tasks. This expert functions as a fallback mechanism, handling inputs that cannot be reliably classified.

\noindent
\textbf{Dynamic Routing.}
While our strategy of manually assigning expert roles allows for fine-tuning with only data corresponding to each type during training, the challenge arises when selecting the appropriate expert in the inference. The most straightforward approach would be to pre-assign a type to the input image, but this increases operational complexity and deviates from practical usage conventions. Therefore, we design a routing mechanism to automatically select the appropriate expert for the input image during inference.
Specifically, the router $r(\cdot)$ is based on an MLP to predict the probability that an image belongs to each data type and selects the expert with the highest score for processing.
The image features extracted by the shared component $g_s$ are used as input for the router.
This process can be formalized as follows, where $\sigma$ denotes $\rm Softmax$:
\begin{equation*}
\mathbf{H_s}=g_s(\mathbf{X_v}),\;
\mathbf{H_v}=e_i(\mathbf{H_s}),\;
i={\rm argmax}\;\sigma(r(\mathbf{H_s})).
\end{equation*}
It is worth noting that many modern MLLMs employ dynamic resolution strategies, where images are split into multiple sub-images. For simplicity, the routing prediction in our approach relies solely on the features of the entire image, with all tokens subjected to global average pooling.

The above describes a na\"ive routing strategy, but the highest-scoring expert may be unreliable for images that are difficult to categorize or have multiple characteristics. To address this, we route the image to the versatile expert designed to handle such challenges. Specifically, we compute the difference $s_d=s_{(1)}-s_{(2)}$ between the highest and second-highest routing scores and compare it with a threshold $\tau$, which can be written as:
\begin{equation*}
\mathbf{H_v} = 
\begin{cases}
e_i(\mathbf{H_s}), & \text{if } s_d \geq \tau \\
e_v(\mathbf{H_s}), & \text{if } s_d < \tau
\end{cases}
\end{equation*}When $s_d$ is small, it suggests that the image type is ambiguous and difficult to classify, in which case we allocate it to the versatile expert to ensure model robustness. More ablations are shown in Sec.~\ref{sec:4_3}.

\section{Experiments}

\subsection{Datasets}
\label{sec:4_1}
% \subsubsection{Datasets}
\textbf{Training Datasets.}
We follow the training datasets used in LLaVA-OV~\cite{li2024llava-ov} and categorize them into five categories based on image characteristics: \emph{Chart}, \emph{Doc}, \emph{Math}, \emph{OCR}, and \emph{General}. The details of our reorganized datasets are shown in Table~\ref{tab:expert_data}. As mentioned in Sec.~\ref{sec:3_3}, different types of datasets will be used to train specialized experts.

\setlength{\tabcolsep}{0.7pt}
\begin{table}
\caption{Details of the reorganized training datasets, all of which are sourced from the LLaVA-OV~\cite{li2024llava-ov} collection.}
\vspace{-10pt}
\renewcommand\arraystretch{1.4}
\begin{center}
\resizebox{\linewidth}{!}{
\begin{tabular}{lllll}
\makecell[l]{\cellcolor[RGB]{70,190,190} \textcolor{white}{\textbf{Chart (0.35M)}}}
& \makecell[l]{\cellcolor[RGB]{73,168,100} \textcolor{white}{\textbf{Math (0.33M)}}}
& \tikz[baseline=0.05em] \fill [color={rgb,255: red,150; green,180; blue,235}] (0,0) rectangle (0.75em,0.75em); K12Printing 
& \tikz[baseline=0.05em] \fill [color={rgb,255: red,225; green,140; blue,135}] (0,0) rectangle (0.75em,0.75em); ShareGPT-4o \\
\tikz[baseline=0.05em] \fill [color={rgb,255: red,230; green,245; blue,245}] (0,0) rectangle (0.75em,0.75em); Chart2Tex 
& \tikz[baseline=0.05em] \fill [color={rgb,255: red,230; green,245; blue,235}] (0,0) rectangle (0.75em,0.75em); MAVIS MCollect 
& \tikz[baseline=0.05em] \fill [color={rgb,255: red,130; green,165; blue,230}] (0,0) rectangle (0.75em,0.75em); OCR-VQA
& \tikz[baseline=0.05em] \fill [color={rgb,255: red,223; green,125; blue,120}] (0,0) rectangle (0.75em,0.75em); ShareGPT4V \\
\tikz[baseline=0.05em] \fill [color={rgb,255: red,180; green,230; blue,230}] (0,0) rectangle (0.75em,0.75em); ChartQA 
& \tikz[baseline=0.05em] \fill [color={rgb,255: red,200; green,235; blue,215}] (0,0) rectangle (0.75em,0.75em); MAVIS Data Engine 
& \tikz[baseline=0.05em] \fill [color={rgb,255: red,110; green,155; blue,225}] (0,0) rectangle (0.75em,0.75em); RenderedText
& \tikz[baseline=0.05em] \fill [color={rgb,255: red,222; green,115; blue,110}] (0,0) rectangle (0.75em,0.75em); ST-VQA \\
\tikz[baseline=0.05em] \fill [color={rgb,255: red,140; green,210; blue,210}] (0,0) rectangle (0.75em,0.75em); DVQA 
& \tikz[baseline=0.05em] \fill [color={rgb,255: red,170; green,220; blue,195}] (0,0) rectangle (0.75em,0.75em); Geo170K QA
& \tikz[baseline=0.05em] \fill [color={rgb,255: red,95; green,145; blue,220}] (0,0) rectangle (0.75em,0.75em); SynthDog-EN
& \tikz[baseline=0.05em] \fill [color={rgb,255: red,221; green,105; blue,100}] (0,0) rectangle (0.75em,0.75em); TallyQA \\
\tikz[baseline=0.05em] \fill [color={rgb,255: red,100; green,200; blue,200}] (0,0) rectangle (0.75em,0.75em); FigureQA
& \tikz[baseline=0.05em] \fill [color={rgb,255: red,140; green,205; blue,175}] (0,0) rectangle (0.75em,0.75em); Geometry3K
& \tikz[baseline=0.05em] \fill [color={rgb,255: red,85; green,140; blue,235}] (0,0) rectangle (0.75em,0.75em); TextCaps 
& \tikz[baseline=0.05em] \fill [color={rgb,255: red,220; green,98; blue,95}] (0,0) rectangle (0.75em,0.75em); VisionFLAN
 \\
\tikz[baseline=0.05em] \fill [color={rgb,255: red,85; green,195; blue,195}] (0,0) rectangle (0.75em,0.75em); Infographic VQA 
& \tikz[baseline=0.05em] \fill [color={rgb,255: red,110; green,190; blue,155}] (0,0) rectangle (0.75em,0.75em); GeoMVerse
& \tikz[baseline=0.05em] \fill [color={rgb,255: red,76; green,137; blue,237}] (0,0) rectangle (0.75em,0.75em); TextOCR
& \tikz[baseline=0.05em] \fill [color={rgb,255: red,220; green,95; blue,90}] (0,0) rectangle (0.75em,0.75em); Visual7W \\
\tikz[baseline=0.05em] \fill [color={rgb,255: red,70; green,190; blue,190}] (0,0) rectangle (0.75em,0.75em);  LRV Chart
& \tikz[baseline=0.05em] \fill [color={rgb,255: red,90; green,180; blue,130}] (0,0) rectangle (0.75em,0.75em); GeoQA+ 
& \makecell[l]{\cellcolor[RGB]{220,91,83} \textcolor{white}{\textbf{General (0.94M)}}}
& \tikz[baseline=0.05em] \fill [color={rgb,255: red,220; green,93; blue,87}] (0,0) rectangle (0.75em,0.75em); VisText \\
\makecell[l]{\cellcolor[RGB]{245,182,53} \textcolor{white}{\textbf{Doc (0.30M)}}} 
& \tikz[baseline=0.05em] \fill [color={rgb,255: red,73; green,168; blue,100}] (0,0) rectangle (0.75em,0.75em); Geo170K Align
& \tikz[baseline=0.05em] \fill [color={rgb,255: red,255; green,235; blue,230}] (0,0) rectangle (0.75em,0.75em); ALLaVA Inst
& \tikz[baseline=0.05em] \fill [color={rgb,255: red,220; green,92; blue,85}] (0,0) rectangle (0.75em,0.75em); VizWiz \\
\tikz[baseline=0.05em] \fill [color={rgb,255: red,255; green,240; blue,200}] (0,0) rectangle (0.75em,0.75em); DocVQA
& \makecell[l]{\cellcolor[RGB]{76,137,237} \textcolor{white}{\textbf{OCR (0.28M)}}}
& \tikz[baseline=0.05em] \fill [color={rgb,255: red,245; green,215; blue,210}] (0,0) rectangle (0.75em,0.75em); AOKVQA
& \tikz[baseline=0.05em] \fill [color={rgb,255: red,220; green,91; blue,84}] (0,0) rectangle (0.75em,0.75em); VQAv2
 \\
\tikz[baseline=0.05em] \fill [color={rgb,255: red,250; green,220; blue,150}] (0,0) rectangle (0.75em,0.75em); RoBUT SQA
& \tikz[baseline=0.05em] \fill [color={rgb,255: red,230; green,240; blue,255}] (0,0) rectangle (0.75em,0.75em); ChromeWriting
& \tikz[baseline=0.05em] \fill [color={rgb,255: red,240; green,200; blue,195}] (0,0) rectangle (0.75em,0.75em); COCO Caption
& \tikz[baseline=0.05em] \fill [color={rgb,255: red,220; green,91; blue,83}] (0,0) rectangle (0.75em,0.75em); WebSight
 \\
\tikz[baseline=0.05em] \fill [color={rgb,255: red,247; green,205; blue,100}] (0,0) rectangle (0.75em,0.75em); RoBUT WikiSQL
& \tikz[baseline=0.05em] \fill [color={rgb,255: red,210; green,225; blue,250}] (0,0) rectangle (0.75em,0.75em); HME100K
& \tikz[baseline=0.05em] \fill [color={rgb,255: red,235; green,185; blue,180}] (0,0) rectangle (0.75em,0.75em); LLaVA-158K
& \\
\tikz[baseline=0.05em] \fill [color={rgb,255: red,246; green,193; blue,75}] (0,0) rectangle (0.75em,0.75em); RoBUTWTQ
& \tikz[baseline=0.05em] \fill [color={rgb,255: red,190; green,210; blue,245}] (0,0) rectangle (0.75em,0.75em); IIIT5K
& \tikz[baseline=0.05em] \fill [color={rgb,255: red,230; green,170; blue,165}] (0,0) rectangle (0.75em,0.75em); LLaVAR
& \\
\tikz[baseline=0.05em] \fill [color={rgb,255: red,245; green,182; blue,53}] (0,0) rectangle (0.75em,0.75em); VisualMRC 
& \tikz[baseline=0.05em] \fill [color={rgb,255: red,170; green,195; blue,240}] (0,0) rectangle (0.75em,0.75em); IAM
& \tikz[baseline=0.05em] \fill [color={rgb,255: red,225; green,155; blue,150}] (0,0) rectangle (0.75em,0.75em); OKVQA 
& \\
\end{tabular}}
\label{tab:expert_data}
\end{center}
\end{table}

\noindent
\textbf{Evaluation Datasets.}
Following previous works~\cite{li2024llava-ov, liu2023visual}, this work primarily focuses on single-image benchmarks. 
Therefore, we have selected several widely-used evaluation datasets within the field, tailored to different task types, including DocVQA test set~\cite{mathew2021docvqa}, ChartQA test set~\cite{masry2022chartqa}, OCRBench test set~\cite{liu2024ocrbenchhiddenmysteryocr}, InfoVQA test set~\cite{mathew2022infographicvqa}, MathVerse mini-vision set~\cite{zhang2025mathverse}, MME test set~\cite{fu2023mme}, AI2D test set~\cite{kembhavi2016diagram}, MathVista testmini set~\cite{lu2023mathvista}, MMBench en-dev set~\cite{liu2025mmbench}.
The above datasets encompass lots of task types faced by multimodal large language models (MLLMs). It is believed that a comprehensive evaluation on these datasets can well reflect the performance of MLLMs.

\noindent
\textbf{Training and Evaluation Datasets for Router.}
We randomly sample 200K entries from each of the five categorized datasets (shown in Table~\ref{tab:expert_data}), constructing a dataset of 1M samples for training the router to assign the appropriate expert for each visual input. 
In addition, we sample 5K chart samples from the ChartQA~\cite{masry2022chartqa} test and validation set, 5K document samples from the DocVQA~\cite{mathew2021docvqa} test set, 5K math samples from the MathVerse~\cite{zhang2025mathverse} and MathVista~\cite{lu2023mathvista} test sets, 5K OCR samples from the IIITK5K~\cite{mishra2012scene} and HME100K~\cite{yuan2022syntax} test sets, 5K general samples from the COCO Caption~\cite{chen2015microsoft} test set, to form a validation dataset consisting of 25K samples for evaluating its accuracy of router in assigning the appropriate expert for each visual input.

\subsection{Implementation Details}\label{sec:4_3}
\noindent
\textbf{Training and Inference Pipeline.}
In summary, our training pipeline comprises expert fine-tuning and router training. Initially, we fine-tuned the experts based on LLaVA-OV~\cite{li2024llava-ov}, with the categorized datasets shown in Table~\ref{tab:expert_data}. During this phase, only the multi-expert layers are trainable, while the shared components remain frozen. Subsequently, we employ the routing dataset described in Sec.~\ref{sec:4_1} to train the router. At the inference phase, we first extract features from the visual input via the shared visual encoder layers. Then, we feed these features into the router, which dynamically chooses the most suitable expert for each visual input based on the routing strategy. 
Furthermore, the selected experts will process the features extracted from the shared layers of the visual encoder and subsequently input these features into Large Language Models.

\noindent
\textbf{Training Details.}
While fine-tuning the added task-specific experts, each expert is trained for one epoch with a global batch size of $256$. A cosine warm-up strategy is employed for adjusting the learning rate, and the maximum learning rate is set as $1\times10^{-5}$ for the projection layer and $2\times10^{-6}$ for the visual encoder. 
We train the router for one epoch with a batch size of $128$, using a cosine annealing schedule, and set the maximum learning rate to $2 \times 10^{-4}$.
AdamW~\cite{loshchilov2017decoupled} serves as the optimizer to train both experts and router, with $\beta1=0.9$, $\beta2=0.98$, and the weight decay of $0.01$, respectively.

\subsection{Ablation Study}\label{sec:4_3}
% ~\ref{sec:4_2}
\textbf{MoE Layer Scanning.} As described in Sec.~\ref{sec:3_3}, this work targets to design an efficient mixture-of-vision-experts architecture, which requires a trade-off between efficiency and performance improvements. To achieve this, we have conducted a comprehensive ablation study to determine which layers are most suitable for being structured as MoE layers. As shown in Table~\ref{tab:moe_layer_scanning}, allowing all layers to be restructured as MoE layers yields the best performance,~\emph{e.g.}, its accuracies on ChartQA and MathVista are 
improved by $1.5\%$ and $1.7\%$ compared to the baseline (LLaVA-OV), respectively.

\setlength{\tabcolsep}{8pt}
\begin{table*}[t]
\renewcommand\arraystretch{1.15}
\caption{\textbf{Ablation study of incorporating different components into the MoE structure.} \# layers indicates the number of vision encoder layers included starting from the deepest (last) layer, where 0 means without the vision encoder, and 27 represents all layers, \ie, the entire vision encoder.
Additional Params refers to the total additional parameters of five experts compared to the baseline (LLaVA-OV). Based on the characteristics of testing samples, we manually selected the specific expert for each evaluation benchmark.}
\vspace{-10pt}
\begin{center}
\resizebox{0.95\textwidth}{!}{
\begin{tabular}{ccc|c|cccccc}
\specialrule{0.1em}{0pt}{0pt}
\multicolumn{3}{c|}{{MoE Layers}} & \multirow{2}{*}{\makecell[c]{{Additional} \\ {Params}}} & {ChartQA} & {DocVQA} & {OCRBench} & {MME} & {MathVerse} & {MathVista} \\ 
 \# layers & Projector & LLM &  & test & test & test & test & mini-vision & testmini \\
\specialrule{0.1em}{0pt}{0pt}
0 & & & 0B & 78.8 & 86.9 & 701 & 2109 & 26.9 & 56.1 \\
27 & \checkmark & \checkmark & 37.41B & 80.3 & 88.4 & 769 & 2183 & 28.6 & 57.8 \\
27 & \checkmark & & 2.14B & 80.0 & 88.1 & 758 & 2166 & 28.2 & 57.6 \\
12 & \checkmark & & 1.00B & 79.9 & 87.9 & 753 & 2157 & 27.9 & 57.3 \\
6 & \checkmark & & 0.54B & 79.7 & 87.8 & 749 & 2151 & 27.7 & 57.1 \\
2 & \checkmark & & 0.24B & 79.5 & 87.6 & 735 & 2146 & 27.5 & 56.8 \\
0 & \checkmark & & 0.09B & 79.1 & 87.3 & 718 & 2128 & 27.2 & 56.5 \\
\specialrule{0.1em}{0pt}{0pt}
\end{tabular}}
\label{tab:moe_layer_scanning}
\end{center}
% \vspace{-6pt}
\end{table*}

\setlength{\tabcolsep}{8pt}
\begin{table}[t]
\caption{\textbf{Ablation studies on the classification accuracy of the router} trained with different scale datasets.}
\label{tab:router_data_scaling}
\vspace{-10pt}
\renewcommand\arraystretch{1.15}
\begin{center}
\resizebox{\columnwidth}{!}{
\begin{tabular}{c|ccccc}
\specialrule{0.1em}{0pt}{0pt}
 Data Size & {\emph{Chart}} & {\emph{Document}} & {\emph{OCR}} &  {\emph{Math}} &  {\emph{General}} \\
\specialrule{0.07em}{0pt}{0pt}
 500K  & 86.4 & 89.8 & 91.6 & 91.0 & 90.4 \\
750K  & 87.9 & 90.6 & 93.9 & 93.2 & 92.5  \\
1M & 89.2 & 91.3 & 95.2 & 94.0 & 93.7 \\   
\specialrule{0.1em}{0pt}{0pt}
\end{tabular}}
\end{center}
\vspace{-6pt}
\end{table}

However, this mechanism significantly increases the overall complexity. Therefore, it is opted to freeze several non-essential layers and adjust the remaining layers into MoE layers. From the Table~\ref{tab:moe_layer_scanning}, it is observed that as fewer layers are restructured as MoE layers, the performances gradually declines. Notably, it is found that the layers best suited for reconstruction are the projector and the deep layers of the Vision Tansformer (ViT)~\cite{vit}. For instance, compared to unfreezing both the projector and all layers of ViT, the mechanism that only modifying the projector and the last two layers of ViT can heavily reduce the additional parameters ($1.9$B) while avoiding substantial performance degradation.
Hence, for the trade-off between efficiency and performance, we simply chose to restructure only the projector and the last two layers of ViT as MoE layers, while keeping the rest of the network unchanged. Additionally, the above modification offers additional advantage of not requiring designing a complex router network. Instead, we can simply use two MLP layers that leverages the features extracted by the shallow layers of ViT to assign the suitable expert for each visual input.

\noindent
\textbf{Performance of Router.}
The performance of router directly affects the overall accuracy of our Mixpert. For instance, if the router incorrectly assigns a mathematics expert for document-type inputs, it could severely degrade the corresponding accuracy. To evaluate whether the designed router is capable of selecting the appropriate expert for visual inputs, we uniformly sampled data of different scales from our organized dataset (as described in Sec.~\ref{sec:4_1}) to train the router and evaluate its classification accuracy for different image types. As shown in Table~\ref{tab:router_data_scaling}, the expert assignment accuracy of the router consistently improves as the training dataset increases. Notably, when training with 1M samples, the router's classification accuracy across various types of datasets is close to or exceeds $90\%$. It achieves $89.2\%$, $91.3\%$, $95.2\%$, $94.0\%$, and $93.7\%$ on the \emph{chart}, \emph{document}, \emph{OCR}, \emph{math}, and \emph{general}, respectively. Moreover, while training the router, only the two added MLP layers are trainable and the shared component always keep frozen. Consequently, the overall training cost is relatively low.

\setlength{\tabcolsep}{4pt}
\begin{table}[t]
\caption{\textbf{Ablation studies of different routing strategies.} ``Direct" refers to selecting the highest-scoring one among five task-specific experts without involving the versatile expert. ``Score-threshold" and ``Score-difference" route difficult-to-categorize images to the versatile expert based on predefined thresholds.}
\label{tab:routing_strategies}
\vspace{-10pt}
\renewcommand\arraystretch{1.15}
\begin{center}
\resizebox{\columnwidth}{!}{
\begin{tabular}{l|ccccc}
\specialrule{0.1em}{0pt}{0pt}
\multirow{2}{*}{Routing} & ChartQA & DocVQA & OCRBench & MME & MathVista\\ 
  & test & test & test & test & testmini \\
\specialrule{0.07em}{0pt}{0pt}
 Direct  & 79.1 & 87.3 & 722 & 2135 & 56.5\\
 Score-threshold  & 79.3 & 87.4 & 728 & 2140 & 56.7\\
 \textbf{Score-difference} & \textbf{79.4} & \textbf{87.5} & \textbf{732} & \textbf{2143} & \textbf{56.7}\\   
\specialrule{0.1em}{0pt}{0pt}
\end{tabular}}
\end{center}
\vspace{-6pt}
\end{table}

\noindent
\textbf{Evaluation on Routing Strategy.}
As mentioned in Sec.~\ref{sec:3_3}, we adopt a score-difference routing strategy. However, there are also some navie routing strategies, such as the direct routing strategy and the score-threshold routing strategy. The direct routing strategy represents that directly selects the most suitable expert among the five experts according to the highest confidence. Additionally, the score-threshold routing strategy means that applies a threshold $\lambda$ to filter the highest confidence among experts and those with highest confidence below the threshold are directly assigned to the versatile expert. 
% In Table.~\ref{tab:routing_strategies}, we conduct an ablation study comparing these three routing strategies. 
% In the ablation study, 
To evaluate the effectiveness of these three routing strategies, we conduct a comprehensive ablation study, in which we use the router trained with 1M dataset and evaluate the above three routing strategies on ChartQA~\cite{masry2022chartqa}, DocVQA~\cite{mathew2021docvqa}, OCRBench~\cite{liu2024ocrbenchhiddenmysteryocr}, MME~\cite{fu2023mme} and MathVista~\cite{lu2023mathvista} benchmark.

As presented in Table~\ref{tab:routing_strategies}, the score-difference routing strategy achieves better performances across various tasks compared to the other two routing mechanisms. Specifically, 
compared to the direct routing, our score-difference strategy can achieve $0.3\%$, $0.2\%$, $10$, $8$, and $0.2\%$ score improvements on {ChartQA~\cite{masry2022chartqa}, DocVQA~\cite{mathew2021docvqa}, OCRBench~\cite{liu2024ocrbenchhiddenmysteryocr}, MME~\cite{fu2023mme}, and MathVista~\cite{lu2023mathvista}, respectively.

Notably, in the above experiments, we set the score-difference threshold $\tau$ as $0.6$. To further verify its influence, we further conduct ablations with different values of $\tau$. As shown in Table~\ref{tab:threshold}, while $\tau$ is set as a small number (\eg, 0.1), our score-difference strategy approximates direct routing and achieves similar results. As $\tau$ increases beyond a certain threshold (\eg, $0.5$), the routing strategy becomes less sensitive to the value changes of $\tau$, and the results across various tasks remains nearly consistent. This observation further demonstrates that the confidence of the assigned expert for each visual input with clear data types is extremely high. As for ambiguous types of visual inputs, setting an appropriate value of $\tau$ can effectively filter out and assign them to the versatile expert, which is the key reason why our score-difference routing strategy outperforms the direct routing mechanism.

\setlength{\tabcolsep}{5pt}
\begin{table}[t]
\caption{\textbf{Ablation studies of different threshold $\tau$} in score-difference routing strategy.}
\vspace{-10pt}
\label{tab:threshold}
\renewcommand\arraystretch{1.15}
\begin{center}
\resizebox{\columnwidth}{!}{
\begin{tabular}{c|ccccc}
\specialrule{0.1em}{0pt}{0pt}
 \multirow{2}{*}{$\tau$} & ChartQA & DocVQA & OCRBench & MME & MathVista\\ 
 % \cmidrule(l){2-6}
  & test & test & test & test  & testmini \\
\specialrule{0.07em}{0pt}{0pt}
 0.1  & 79.2 & 87.3 & 725 & 2132 & 56.5\\
 0.3  & 79.2 & 87.4 & 727 & 2133 & 56.5\\
 0.5  & 79.3 & 87.5 & 732 & 2142 &  56.7\\
 \textbf{0.6} & \textbf{79.4} & \textbf{87.5} & \textbf{732} & \textbf{2143} & \textbf{56.7} \\ 
 0.7 & 79.4 & 87.4 & 731 & 2143 & 56.5\\ 
\specialrule{0.1em}{0pt}{0pt}
\end{tabular}}
\end{center}
\vspace{-6pt}
\end{table}

\noindent
\textbf{Additional Cost.}
Once the overall architecture of Mixpert has been confirmed, the remaining question is how much additional complexity Mixpert introduces compared to the baseline (LLaVA-OV~\cite{li2024llava-ov}). To address this, we calculate the total parameters, activated parameters, and computation cost for both LLaVA-OV and Mixpert in details. As shown in Table~\ref{tab:additional_cost}, Mixpert adds only an additional 237.1M parameters compared to LLaVA-OV. 
Since Mixpert only selects the most suitable expert for each visual input, its additional activated parameters and computation cost during inference phrase are primarily due to the router module, amounting to 1.3M additional parameters and {0.001G} FLOPs, respectively. Compared to the overall parameters and computation overhead of LLaVA-OV, these increments are negligible.

\noindent
\textbf{The Benefits from Fine-tuning.} Another issue arises from the fact that Mixpert requires additional fine-tuning for each expert. Consequently, it is pertinent to inquire whether directly fine-tune the last two vision layers and the projector would also yield additional improvements. To evaluate this, we further conduct fine-tuning on LLaVA-OV and other state-of-the-arts with all the collected expert data, respectively.
As shown in Table~\ref{tab:additional}, the performance gains of Mixpert primarily stems from its appropriate dynamic routing scheme rather than the additional fine-tuning process.

\setlength{\tabcolsep}{3pt}
\begin{table}[t]
\caption{\textbf{Comparisons of parameters and computation costs} between the baseline (LLaVA-OV-7B) and ours.}
\vspace{-10pt}
\label{tab:additional_cost}
\renewcommand\arraystretch{1.2}
\begin{center}
\resizebox{\columnwidth}{!}{
\begin{tabular}{c|ccc}
\specialrule{0.1em}{0pt}{0pt}
 Method & Params & Activated Params & FLOPS  \\ 
\specialrule{0.07em}{0pt}{0pt}
 LLaVA-OV-7B   & 7482.1M & 7482.1M &  5451.157G \\
\name (LLaVA-OV-7B)   & 7719.2M & 7483.4M &  5451.158G \\
\specialrule{0.1em}{0pt}{0pt}
\end{tabular}}
\end{center}
\end{table}

\setlength{\tabcolsep}{1pt}
\begin{table}[t]
\caption{\textbf{Additional training.}  *: the results achieved by further fine-tuning the last two layers of the visual
encoder and the projector.}
\vspace{-10pt}
\label{tab:additional}
\renewcommand\arraystretch{1}
\begin{center}
\resizebox{\columnwidth}{!}{
\begin{tabular}{c|ccccc}
\specialrule{0.1em}{0pt}{0pt}
  \multirow{2}{*}{Model} & ChartQA & DocVQA & OCRBench & MME & MathVista\\ 
 % \cmidrule(l){2-6}
  & test & test & test & test  & testmini \\
 
\specialrule{0.05em}{0pt}{0pt}
LLaVA-OV-7B* & 78.7 & 86.7 & 694 & 2090 & 56.3 \\
Mixpert(LLaVA-OV-7B) & \textbf{79.4} & \textbf{87.5} & \textbf{732} & \textbf{2143} & \textbf{56.7} \\
\specialrule{0.05em}{0pt}{0pt}
InternVL2-8B* & 83.0 & 91.4 & 789 & 2207 & 57.9 \\ 
Mixpert(InternVL2-8B) & \textbf{84.0} & \textbf{92.3} & \textbf{806} & \textbf{2271} & \textbf{58.8} \\
\specialrule{0.1em}{0pt}{0pt}
\end{tabular}}
\end{center}
\end{table}

\subsection{Comparisons with State-of-the-Arts}
\textbf{Comparisons with Fully Open-Source Methods.} As a well-known open-source project in the field of MLLMs, LLaVA series~\cite{liu2023visual,li2024llava-ov,liu2024llava} have fully released both their training data and well-trained model weights. Therefore, the proposed Mixpert is primarily built upon LLaVA-OV~\cite{li2024llava-ov} to conduct comprehensive and fair comparisons on various multimodal benchmarks. As described in Sec.~\ref{sec:4_1}, the training of all newly added experts and the router in Mixpert is carried out using the same data corpus utilized by LLaVA-OV. Thus, the improvements of Mixpert over LLaVA-OV are primarily due to the methodology itself, rather than new training datasets.

\setlength{\tabcolsep}{5pt}
\begin{table*}[t]
\renewcommand\arraystretch{1.15}
\caption{\textbf{Comparisons between \name and other state-of-the-arts} across different types of commonly used benchmarks. *: the evaluation using chain-of-thought prompting; \dag: the results tested by ourselves with official checkpoints; MI: the multi-image version of LLaVA-OV; \S: the results achieved by the sub-image routing strategy rather than the global-image routing strategy.}
\vspace{-10pt}
\begin{center}
\resizebox{\textwidth}{!}{
\begin{tabular}{l|ccccccccc}
\specialrule{0.1em}{0pt}{1pt}
 \multirow{2}{*}{{Model}}  & {ChartQA} & {DocVQA} & {OCRBench} & {MME} & {AI2D} & {InfoVQA} & {MathVerse} & {MathVista} & {MMBench} \\ 
 % \cmidrule(l){2-10}
  &  test & test & test & test & test & test & mini-vision & testmini & en-dev \\
\specialrule{0.1em}{0pt}{0pt}
\rowcolor{Gray!25}
\emph{\textbf{Open-source}} & & & & & & & & & \\
Cambrian-34B~\cite{tong2024cambrian} & 75.6 & 75.5 & - & - & 79.7 & - & - & 53.2 & 81.4 \\
% VILA-34B\cite{lin2024vila} & - & - & - & 1762 & - & - & - & - & 82.4 \\
Eagle-X5-13B~\cite{shi2024eagle} & 71.0 & - & 573 & 1604 & - & - & - & 39.7 & 70.5 \\
SPHINX-MoE~\cite{liu2024sphinx} & 55.0 & 68.4 & - & 1852 & 55.6 & 41.8 & - & - & 71.3 \\
Mini-Gemini-35B~\cite{li2024mini} & - & - & - & 2141 & - & - & - & 43.3 & 80.6 \\
CuMo (Mistral-8x7B)~\cite{li2024cumo} & - & - & - & 1640 & - & - & - & 38.2 & 75.3 \\
LLaVA-OV-7B~\cite{li2024llava-ov} & 78.8 & 86.9 & 701 & 2109 & 81.6 & 65.3 & 26.9 & 56.1 & 81.7\\

LLaVA-OV-7B(MI)~\cite{li2024llava-ov} & 80.0 & 87.5 & 621 & 1998 & 81.4 & 68.8 & 26.2 & 63.2 & 80.8\\
 \specialrule{0.05em}{0pt}{0pt}
\rowcolor{Blue!10}
\textbf{\name (LLaVA-OV-7B)} & \textbf{79.4} & \textbf{87.5} & \textbf{732} & \textbf{2143} & \textbf{82.0} & \textbf{65.9} & \textbf{27.4} & \textbf{56.7} & \textbf{82.2}\\
\rowcolor{Blue!10}
\textbf{\name (LLaVA-OV-7B(MI))} & \textbf{81.1} & \textbf{88.3} & \textbf{659} & \textbf{2021} & \textbf{81.8} & \textbf{69.4} & \textbf{27.0} & \textbf{63.8} & \textbf{81.5}\\
\rowcolor{Blue!10}
\textbf{\name (LLaVA-OV-7B(MI))$^\S$} & \textbf{81.4} & \textbf{88.6} & \textbf{667} & \textbf{2030} & \textbf{82.0} & \textbf{69.8} & \textbf{27.1} & \textbf{64.0} & \textbf{81.9}\\
 \specialrule{0.07em}{0pt}{2pt}
 \specialrule{0.07em}{0pt}{0pt}
 \rowcolor{Gray!25}
\emph{\textbf{Open-weights}} &  & & & & & & & & \\

MiniCPM-V2.6~\cite{yao2024minicpm} & - & 90.8 & 852* & 2348* & 82.1 & - & - & 60.6 & - \\

InternVL2-8B~\cite{chen2024far} & 83.3 & 91.6 & 794 & 2210 & 83.8 & 74.8 & 27.5 & 58.3 & 81.7\\
Qwen2-VL-7B$^{\dag}$~\cite{wang2024qwen2} & 82.9 & 94.4 & 863 & 2328 & 82.8 & 76.6 & 25.8 & 58.1 & 82.9\\
 \specialrule{0.05em}{0pt}{0pt}
 \rowcolor{Blue!10}
\textbf{\name (InternVL2-8B)} & \textbf{84.0} & \textbf{92.3} & \textbf{806} & \textbf{2271} & \textbf{84.1} & \textbf{75.3} & \textbf{27.9}  & \textbf{58.8} & \textbf{82.3}\\
\rowcolor{Blue!10}
\textbf{\name (Qwen2-VL-7B)} & \textbf{83.4} & \textbf{94.8} &  \textbf{873} & \textbf{2346} & \textbf{83.2} & \textbf{76.9} & \textbf{26.4} & \textbf{58.4} & \textbf{83.4} \\
\specialrule{0.1em}{0pt}{0pt}
\end{tabular}}
\label{tab:results}
\end{center}
% \vspace{-6pt}
\end{table*}

As shown in Table~\ref{tab:results}, Mixpert demonstrates varying degrees of improvement over LLaVA-OV~\cite{li2024llava-ov} across multiple multimodal benchmarks. Specifically, there is a $0.6\%$ improvement compared to LLaVA-OV-7B~\cite{li2024llava-ov} on both ChartQA~\cite{masry2022chartqa} and DocVQA~\cite{mathew2021docvqa}. In the OCRBench~\cite{liu2024ocrbenchhiddenmysteryocr} and MME~\cite{fu2023mme}, \name achieves scores of $732$ and $2143$, surpasses LLaVA-OV-7B~\cite{li2024llava-ov} with $31$ and $34$, respectively. 
In other benchmarks, such as MathVerse~\cite{zhang2025mathverse} and MathVista~\cite{lu2023mathvista}, \name has also achieved clear improvements. 
Furthermore, LLaVA-OV(MI) is the version trained on multi-image datasets. Considering that each image or sub-image splitted from a single image may belong to distinct categories, we also conduct experiments with independent routing strategy for each sub-image. As shown in Table~\ref{tab:results}, indeed, compared to the global-image routing strategy, the sub-image routing strategy brings more improvements. Of course, adhering to the sub-image routing strategy also introduces additional costs, specifically increasing the activated parameters for each test image. This is the reason why Mixpert directly opts for global-image routing.
Additionally, when compared to recent methods that incorporating multiple vision encoders or designing MoE architectures (\emph{e.g.},  EAGLE~\cite{shi2024eagle} and CuMo~\cite{li2024cumo}), \name still shows competitive results across various tasks.

\noindent
\textbf{Comparisons with Open-Weights Methods.}
Many state-of-the-art methods have released model weights, yet not training data. 
To further validate the effectiveness of \name, we integrate it into two recent well-known MLLMs: InternVL2-8B~\cite{chen2024far} and Qwen2-VL-7B~\cite{wang2024qwen2}. Notably, the visual encoder used in InternVL2-8B~\cite{chen2024far} is InternViT~\cite{chen2024internvl}, whereas Qwen2-VL-7B~\cite{wang2024qwen2} employs a self-designed ViT architecture with support for naive dynamic resolution of inputs. However, for both InternVL2-8B~\cite{chen2024far} and Qwen2-VL-7B~\cite{wang2024qwen2}, we only restructure the corresponding projector module and the last two layers of the visual encoder into MoE layers. We collect approximately 5.5M samples from publicly available datasets, which are simply categorized into three types: \emph{chart}, \emph{document}, and \emph{general}. As a result, Mixpert includes four experts: versatile, chart, document, and general experts. Additionally, 3M images are randomly sampled from the above-collected datasets and used to train the router.

As indicated by the results in Table~\ref{tab:results}, even built upon state-of-the-art methods that are trained on unknown training data, Mixpert is still able to improve the performance across various benchmarks.
Compared to InternVL2-8B~\cite{chen2024far}, Mixpert demonstrates a significant improvement of $0.7\%$, $0.7\%$, and $61$ on ChartQA~\cite{masry2022chartqa}, DocVQA~\cite{mathew2021docvqa}, and MME~\cite{fu2023mme}, respectively. 
Even evaluated on the currently most powerful open-weights model, Qwen2-VL-7B~\cite{wang2024qwen2}, Mixpert still achieves consistent improvements.
The above results further demonstrate that Mixpert can integrate seamlessly into any MLLM, effectively mitigating the underlying multi-task learning conflicts, which further enhances the visual perception capabilities of MLLMs while handling different types of visual inputs.

\section{Conclusion}
This work firstly reveals the multi-task learning conflicts within a single visual encoder that are commonly faced by current multimodal large language models (MLLMs), and then points out that recent works aiming to enhance visual perception capabilities by directly integrating multiple visual encoders will inevitably add heavy computation overhead. To address this, this paper further proposes Mixpert, an efficient mixture-of-vision-experts architecture to assign the most suitable domain-specific expert for each visual input. Comprehensive experiments have substantiated the effectiveness of Mixpert, which can seamlessly integrate into any MLLM and significantly improve the corresponding performances on various scenarios with less additional cost. We believe that Mixpert is an initial attempt to address the multi-task learning conflicts while minimizing the additional complexity in MLLMs, and more efficient approaches are worth exploring in future work.
\clearpage

{
    \small
    \bibliographystyle{ieeenat_fullname}
    \bibliography{main}
}

\end{document}